\documentclass[letterpaper, 10 pt, conference]{ieeeconf}  

\IEEEoverridecommandlockouts                              
\usepackage{graphicx} 
\usepackage{caption}
\usepackage{subcaption}
\usepackage{amsmath}
\usepackage{amsfonts}
\usepackage{mathrsfs}  
\usepackage{enumerate}
\usepackage{algpseudocode}
\usepackage[ruled,vlined]{algorithm2e}                                                       

\usepackage{amsmath}
\DeclareMathOperator{\Tr}{Tr}

\DeclareMathOperator*{\argmin}{\arg\!\min}
\title{\LARGE \bf
A Sensorimotor Reinforcement Learning Framework \\ for Physical Human-Robot Interaction 
}

\author{Ali Ghadirzadeh,  Judith B\"utepage, Atsuto Maki, Danica Kragic and M{\r a}rten Bj\"orkman
\thanks{Authors are with the Computer Vision and Active Perception Lab (CVAP), CSC, KTH Royal Institute of Technology, Stockholm, Sweden.
        {\tt\small algh|butepage|atsuto|dani|celle@kth.se}}%
}

\begin{document}

\maketitle
\thispagestyle{empty}
\pagestyle{empty}

\begin{abstract}

Modeling of physical human-robot collaborations is generally a challenging problem due to the unpredictive nature of human behavior. 

To address this issue, we present a data-efficient reinforcement learning framework which enables a robot to learn how to collaborate with a human partner. 
The robot learns the task from its own sensorimotor experiences in an unsupervised manner. 

The uncertainty of the human actions is modeled using Gaussian processes (GP) to implement action-value functions.
Optimal action selection given the uncertain GP model is ensured by Bayesian optimization.  

We apply the framework to a scenario in which a human and a PR2 robot jointly control the ball position on a plank based on vision and force/torque data. 
Our experimental results show the suitability of the proposed method in terms of fast and data-efficient model learning, optimal action selection under uncertainties and equal role sharing between the partners.  
\end{abstract}
\section{INTRODUCTION}
\label{sec:intro}

As we envision robots to closely collaborate with humans in a growing number of applications, the study of physical Human-Robot Interaction (pHRI) is gaining in importance. The most crucial aspect of pHRI is to ensure safety, but the robot should also act intuitively for the human partner.
A robot, commonly in physical contact with a human collaborator, should not only be proactive in responding to the human's intentions, but also actively contribute to achieve the shared goal.

\begin{figure}[t]
\begin{center}
\includegraphics[width=0.48\textwidth ]{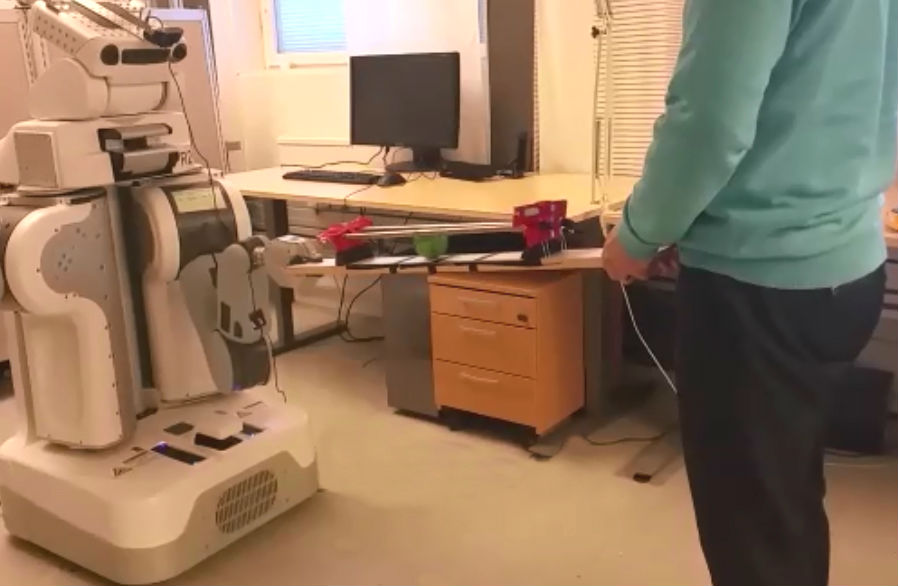}
\caption{The collaborative task setup. The aim is to move the green ball on the jointly held plank to a predefined goal position. Both the human and the PR2 robot should actively contribute to reach the shared goal.}
\vspace{-8 mm}
\label{fig:setup}
\end{center}
\end{figure}

In this work, we consider the problem of physical human-robot interaction in a dynamic control task based on sensorimotor reinforcement learning. The task is to control the position of a ball on a jointly held  plank using vision and force/torque data, as shown in Fig.~\ref{fig:setup}. 
We assume that there is no predefined leader/follower role assignments for the partners and both the human and the robot have equal roles in the task. 

Why are pHRI tasks, such as the one described above, generally so challenging? 
The main difficulty lies in the problem of modeling and predicting human behavior.
Although some low-level models of human motions exist, e.g.\ the minimum jerk model \cite{maeda2001human}, only few studies deal with long-term human motion prediction or finding kinematic constrains imposed by the human body (e.g. \cite{karayiannidis2013online}). 
pHRI is a demanding task due the lack of proper models, the unpredictability of  human behavior and the dynamic nature of the problem.

To deal with these difficulties, many studies within pHRI have introduced the robot as a passive partner that follows the action plan of a human leader. 
This is not always a suitable formulation. First of all, conveying the human's command to a robot can be challenging (see e.g. \cite{dumora2012experimental} and \cite{karayiannidis2014mapping}). Furthermore, the structure of the task might hinder one-sided decision making and require both partners to actively contribute.

In a broader context, pHRI can be formulated as (i) human-leader and robot-follower, (ii) human-follower and robot-leader or (iii) equal role sharing. 
In the first case, the robot can be a proactive follower that reacts to the estimated human motion profile while the human is responsible for the motion planning and conveying appropriate signals to the robot. 
In the second case, the robot carries the responsibility for task completion while the human is assumed to be a compliant cooperator.
Both of these two role assignments might decrease the efficiency of what the collaborators can do together and limit the applicability of the method. 
On the other hand, in an equal role sharing scenario, both partners can collaboratively support each other. Especially in cases that one partner cannot fully contribute due to a temporary limited observability. In this case, the equal role sharing results in a smoother interaction. 

An important element for the robot to contribute equally is to have a  proper model representing human interactions while being physically connected to the robot. 
For a smooth interaction, it is important for each agent to predict the partner and incorporate its action-state while making decisions. 
Humans are naturally equipped with such a capability, however it is challenging to fit similar ability in robots. 
Several studies (e.g.\ \cite{bussy2012proactive}, \cite{agravante2013human} and \cite{agravante2014collaborative}) modeled human actions as disturbances. 
Others (e.g.\ \cite{thobbi2011using}, \cite{bussy2012proactive} and \cite{gribovskaya2011motion}) predict future actions of the human and proactively generate motion plans that are compliant with this estimate. 
A few studies assume a simplified model, e.g. a revolute joint, instead of an accurate representation \cite{karayiannidis2013online}. 

We propose a sensorimotor approach toward pHRI.
The task is grounded directly in the sensorimotor space of the robot. This avoids high-level symbolic representation of the task and the human. Instead, the interaction model is learned by an adequate number of active observations, i.e. performing actions and observing the consequences. This mapping between sensory states, actions and their effects is termed sensorimotor contingencies (SMCs) \cite{o2001like}.


It is generally difficult to hand-design an optimal action-selection policy based on SMCs maps. Such maps are usually non-linear and require further preprocessing. Alternatively, reinforcement learning (RL) methods are well-suited to find optimal policies even for high-dimensional sensorimotor spaces.
We apply probabilistic model-based Q-learning, a RL technique, to find optimal actions that  optimize the expected return of an objective function (see \cite{kober2012reinforcement} for an in-depth discussion). 
In the next section, we summarize related work in pHRI domain.

\subsection{Related work}
Traditionally, studies in pHRI assume a fixed role distribution between robotic and human partners. In most cases, the robot is viewed as the follower while the human partner takes the leading responsibility.
One example of this role assignment was proposed by 
Wojtara et al.\ \cite{wojtara2009human} for cooperative positioning of jointly manipulated objects. The human partner determines the manipulation trajectory while the robot follows the human commands and compensates for the load. 
Dumora et al.\ \cite{dumora2012experimental} and Karayiannidis et al.\ \cite{karayiannidis2014mapping} investigated how to disambiguate between translational and rotational human commands in scenarios in which an object is jointly manipulated with a human partner. 
These methods are mostly suited for industrial assembly of heavy objects. 
On the other extreme, some studies consider the robot as the leader in a cooperative task. As an example, in the work presented by Karayiannidis et al.\ \cite{karayiannidis2013online} the kinematic model of the human partner is represented as a passive revolute joint. In their experimental results, they  show how this model can be used to keep a jointly held object at a horizontal position, while the robot is leading the interaction. 
In comparison to these studies, we aim for a more collaborative role assignment between the partners and an equal responsibility to complete the task. 

To involve the robot more in a cooperative task, proactive behavior has been proposed. This term implies that the robot should proactively adjust its motion profile to facilitate the action trajectory devised by the human. For this purpose, the robot is equipped or trained with a model of task-relevant human behavior. By means of such a model, the robot is able to predict human actions and incorporate this knowledge into its own action planning. 
Maeda et al.\ \cite{maeda2001human} suggested to predict human motion profiles with the help of minimum jerk model and to implement proactive action selection as a variant of impedance controller. They reported that this method makes the cooperative beam-lifting task more comfortable for the human operators.  A major limitation of the minimum jerk model is that the desired final configuration  needs to be known a priori.
In a similar approach, Thobbi et al.\ \cite{thobbi2011using} modeled human motion profile in a proactive table-lifting task. Instead of the minimum jerk model, the authors used an extended Kalman filter that does not require a predefined goal position. The robot combines reactive and proactive behaviors based on how confident it is about predicting the next action of the human partner.
An adaptive impedance controller has been proposed by Gribovskaya et al.\ \cite{gribovskaya2011motion} for a cooperative beam-lifting task. The framework is based on learning by demonstration. The robot is controlled by a human user to observe a number of cooperative trajectories encoded as the end-effector's velocities and the interaction forces. Subsequently, a dynamic forward model is trained using the demonstrated trajectories to approximate state evolution. 
Finally, the estimated states are used as the reference point for the impedance controller. 
Although a proactive behavior makes the interaction more intuitive for the human partner, the robot is still viewed as a follower. 

Role switching, i.e.\ to switch between leader and follower roles conditioned on e.g.\ limited observability or uncertainty, is investigated in a number of studies.
As an example, Ervard and Kheddar \cite{evrard2009homotopy} introduced a homotopy switching model. The robot behavior is governed by two extreme leader and follower actions. For the joint object manipulation scenario, the leader action compensates for the positioning error while the follower action minimizes the interaction force sensed at the grasp point. The final output is a weighted average of these two actions. 
However, it is not clear how these weights should be adjusted during the task. 
In our work, we do not assume an explicit switching mechanism. 
Instead, both the positioning error and the interaction force are optimized simultaneously by optimizing an objective function.  

Equal role sharing is accomplished in a number of studies using impedance control. Impedance control enables the robot to impose its own motion trajectory while being flexible to the partner's one.  
Bussy et al.\ \cite{bussy2012proactive} suggested an impedance control law that determines a reference trajectory for the controller to follow a desired trajectory. They claimed that both leader and follower roles can be realized using the same controller type by providing appropriate desired trajectories. In a joint table-carrying task, the robot predicts the human motion plan by matching a set of motion primitives. Subsequently, it generates a trajectory that is consistent with the prediction. To realize a leader role, the task is repeated with the robot being teleoperated by another human user. 
Similarly, Agravante et al.\ \cite{agravante2013human} used the same method and setup with the constraint that the table should not be tilted. The tilt angle of the table is measured by processing camera images and a desired trajectory is generated by a visual servoing system instead of teleoperation. 
In their more recent work \cite{agravante2014collaborative}, they modified the task such that the robot should also prevent objects on the table from falling down.
Although we aim to solve a similar problem, i.e.\ formulating a collaborative pHRI with equal roles, our approach is different. 
Impedance control methods require well-calibrated sensory measurements as well as a proper dynamic model of the system. 
Furthermore, a set of impedance parameters has to be found and tuned. Unlike this, we use a generic learning framework which does not require human expertise to design the system. 

In this work, we present a sensorimotor reinforcement learning framework. Uncertainty in the pHRI task is addressed by Bayesian modeling and optimization. This results in a model that not only gives how confident it is in making predictions, but also generates an optimal policy even under uncertainty assumptions. 
Equal role sharing is achieved by devising an objective function for the robot similar to the one of the human partner. Both agents mutually model and predict their partners to select actions that accomplish the task goal while maintaining a smooth interaction. 

\section{METHOD}
In this section, we present the details of our proposed method. 
The framework consists of two components: a forward model and an action-value function (Q-function). 
The forward model predicts sensory outcomes of the robot's actions, while the Q-function represents expected accumulated cost for an action-state pair (note that, unlike standard RL problems, we model \textit{cost} instead of \textit{reward}). 
The forward model provides training samples to construct the Q-function and the Q-function, once it is constructed, is used to find the optimal policy. 

Such a model-based approach can result in an efficient use of the data collected from the robot (see e.g.\ \cite{deisenroth15} and \cite{deisenroth2013survey}).  
Since constructing utility functions generally might require many training samples over trials that, in the case of a model-based method, the data can be provided by the forward model, instead. 

The structure of this section is as follows:
In Sec.~\ref{sec:representation}, we introduce the sensorimotor representation of the system; Sec.~\ref{sec:forward} and ~\ref{sec:GP} describe forward model learning using Gaussian processes and Sec.~\ref{sec:GPQ} introduces the probabilistic model-based version of Q-learning algorithm. 

\begin{figure}[b!]
\begin{center}
\includegraphics[width=0.35\textwidth]{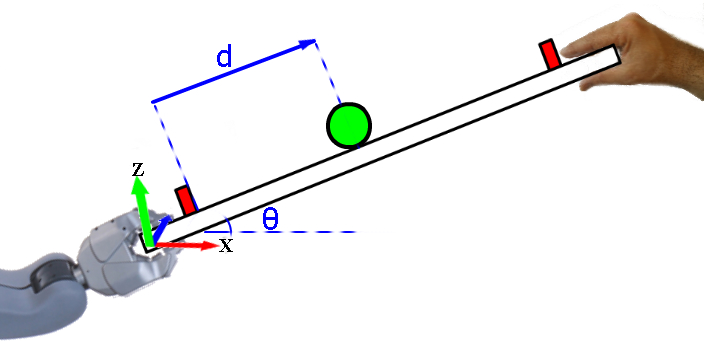}
\caption{The ball position $d$ and velocity $\dot{d}$ are encoded in the plank reference frame. The robot is controlling velocities in $X-Z$ plane and the pitch angular velocity $\dot{\theta}$ in the end-effector reference frame.}
\label{fig:robot}
\end{center}
\end{figure}

\subsection{Sensorimotor representation}
\label{sec:representation}
The sensory observations are denoted as a state vector 
 $s_t = [s_1, ..., s_{n_s}]$ and motor commands as an action vector $a_t=[a_1, ..., a_{n_a}]$ at time step $t$.
The actions are velocity commands sent to the Cartesian velocity controller with the frame at the robot's end-effector. 
As shown in Fig.~\ref{fig:robot}, the $X-Z$ plane velocities as well as pitch angular velocity are used to roll the ball on top of the plank. 
Thus, the action space is represented by the command velocities $a_t= [\dot{x}_t,\dot{z}_t, \dot{\theta}_t$]. 
The velocity in $y$ direction as well as roll and yaw angular velocities are controlled to fix the plank at $y = 0$, roll angle $= \pi/2$ and yaw angle $= 0$. 

The state at time $t$ consists of the end-effector pose [$x_t, y_t, \theta_t$] and the interaction force $\tau_t$, calculated as the sum of absolute uncalibrated torques in all the three dimensions. Additionally, the ball position and velocity [$d_t, \dot{d}_t$] and the distance to the current ball goal position $\Delta d_t = d_{t} - d_t^*$, all measured by the vision system, are included in the state vector. In summary, an input state is defined as $s_t= [x_t,z_t,\theta_t,\Delta d_t, d_t, \dot{d}_t,\tau_t$]. 

The aim of the task is to control the ball to stay at the target position with a minimum interaction force; Therefore the cost is defined as $c(s_t) =[\Delta d_t, \dot{d_t}, \tau_t]W_c[\Delta d_t, \dot{d_t}, \tau_t]^T$, where $W_c$ is a diagonal weight matrix. 

\subsection{Forward model learning}
\label{sec:forward}
As mentioned earlier, the forward model predicts how the states will be affected as the consequence of the robot's actions. Let $\Delta s_t = s_{t+1} - s_t$ describes the state change induced by the action $a_t$. Then a forward model predicts these changes as below:
\begin{equation} 
	\Delta s_t = \mathscr{F} (s_t , a_t).
	\label{eqForward}
\end{equation}
Following our previous work \cite{Ghadirzadeh14}, \cite{Ghadirzadeh15} and \cite{ghadirzadeh2016self}, we apply Gaussian Processes (GP) to implement the forward model. 
In the last two studies, we formulated a data-efficient, incrementally trained sensorimotor model based on GPs that will be briefly introduced in the following section.

\subsection{Gaussian process regression} 
\label{sec:GP}
Gaussian processes are used to implement the Q-function and each dimension of the forward model. 
Totally $n_s + 1$ GP models are trained independent of each other. 
The training samples for each GP are given by a set of $N$ state-action vectors $\mathbf{X}=(s_i,a_i)_{i = 1,...,N}$ 
as the input, and a column-vector $Y$ as the target values. 
$Y$ consists of the changes in the corresponding state for training the forward model and the Q-values for the Q-function learning. 

The GP prior mean function is chosen to be zero ($m(x) = 0$) and the squared exponential function
\begin{equation}
k(x,x') = \sigma_f^2 exp(-\frac{1}{2}(x-x')^TW^{-1}(x-x')),
\label{eq:covSEard}
\end{equation}
is chosen as the GP prior covariance function. 
$\sigma_f$ and $W$ are the hyperparameters found by minimizing the marginal log-likelihood of the training data.  

For a test data point $x_t=(s_t,a_t)$, the test-train and train-train co-variances are denoted by the vector $k(x_t, \mathbf{X})$ and the matrix $\mathbf{K} = k(\mathbf{X},\mathbf{X})$, respectively.

The GP regression output for an input point $x_t$, is given by the posterior mean
\begin{equation} 
	\label{eq:GP_Mean}
	m_* = k(x_t,\mathbf{X})\beta,
\end{equation}  
and the posterior variance
\begin{equation} 
	\label{eq:GP_Variance}
	v_* = k(x_t,x_t) - k(x_t,\mathbf{X}) (\mathbf{K}+\sigma_n \mathbf{I})^{-1} k(x_t,\mathbf{X}),
\end{equation} 
where $\beta = (\mathbf{K}+\sigma_n \mathbf{I})^{-1}Y$ 
and $\sigma_n$ represents the measurement noise that is found the same way as the other hyperparameters. 

\subsection{Probabilistic model-based Q-learning}
\label{sec:GPQ}
The Q-function $Q(s,a)$ represents the expected return of being in state $s$ and taking action $a$ while following the optimal policy afterwards.
It is updated incrementally for the state-action pair $(s_t,a_t)$ based on the predicted next state $s_{t+1} \sim \mathcal{N}(\mu_{t+1}, \Sigma_{t+1})$
\begin{equation} 
	Q(s_t,a_t)  \leftarrow \mathbb{E}_{s \sim s_{t+1}}[c(s)] + \gamma \min_{a'} \mathbb{E}_{s\sim s_{t+1}}[Q(s,a')] ,
	\label{eq:Qupdate}
\end{equation}
where $\gamma$ is the discount factor and
\begin{equation} 
\mathbb{E}_{s\sim s_{t+1}}[c(s)] = \int p(s' | \mu_{t+1} , \Sigma_{t+1}) c(s') ds'.
\end{equation}
The cost is calculated as the weighted squared Euclidean distance to the target state $s^*$. 
Substituting $c(s) = (s - s^*)W_c (s -  s^*)^T$ in the above relation yields
\begin{equation} 
\begin{split}
\mathbb{E}_{s\sim s_{t+1}}[c(s)] = (\mu_{t+1} - s^*)\,W_c\,(\mu_{t+1} - s^*)^T \\ + \, \Tr \,(W_c\,\Sigma_{t+1}).
\end{split}
\end{equation} 

The Q-function is a zero-mean Gaussian process $(Q \sim \mathcal{GP}(m(x) = 0, k))$ with the covariance function introduced in Eq. \ref{eq:covSEard}, the hyperparameters $\Theta_q = [W, \sigma_f, \sigma_n]$ and the training data $\mathcal{D} = [\mathbf{X}, Y]$. 
The predictive distribution of the Q-function at a test action-state point $x_t = [s_t, a_t]$ is a normal 
\begin{equation} 
p(q|x_t, \theta_q) = \mathcal{N}(m_*(x_t), k_*(x_t, x_t)),
\end{equation}
where, $m_*$ and $k_*$ are the GP posterior mean and covariance functions, respectively. 
The expected Q-value over the predicted next state is represented by:
\begin{equation} 
	\mathbb{E}_{s\sim s_{t+1}}[Q(s,a)] = \iint p(q|s', a, \theta_q) p(s' | \mu_{t+1} , \Sigma_{t+1})q \,ds'dq, 
\end{equation}
and can be found analytically according to \cite{kuss2006gaussian}:
\begin{equation} 
\mathbb{E}_{s\sim s_{t+1}}[Q(s,a)] = |\Sigma_{t+1}W^{-1} + I|^{-\frac{1}{2}}\beta L,
\end{equation}
where $L$ is a vector defined as:
\begin{equation} 
L = \sigma_s^2 exp(-\frac{1}{2}(\mathbf{X}-\mu_{t+1})(\Sigma_{t+1}+W)^{-1}(\mathbf{X} -\mu_{t+1})^T).
\end{equation}
In the following subsections, we introduce action-selection method and how to incrementally train the Q-function based on forward model simulations.
\subsubsection{Action selection} 
\label{sec:act_bay}
Once the Q-function is constructed, the optimal policy  chooses the action that minimizes the Q-function for a given state. 
However, due to the limited number of training samples, as well as stochasticity of the pHRI system, the uncertainty of the Q-function should be taken into account.  
Generally, such uncertainties are avoided by limiting state-action trajectories to be close to the previous ones. For example, the KL-divergence measure is used as the constraint while optimizing trajectories \cite{schulman2015trust}. 
Here, we implement Bayesian optimization \cite{shahriari2016taking} that is a standard method to deal with action planning under uncertainty assumptions. Instead of constraining the action-state trajectory, we penalize uncertainty according to the Upper Confidence Bound (UCB)
\begin{equation} 
Q_{UCB}(x_t) = m_*(x_t)  - \delta \sigma_*(x_t), 
\label{eq:costPi}
\end{equation} 
where $x_t = [s_t, a_t]$, $m_*(.)$ and $\sigma_*(.) = [k_*(.,.)]^\frac{1}{2}$ are the GP posterior mean and standard deviation and $\delta$ is a weighting factor.
And the optimal action minimizes $Q_{UCB}$
\begin{equation}
a_{t+1} = \argmin_{a} Q_{UCB}(s_t, a)
\label{eq:policy}
\end{equation}
We use Gradient based optimization methods to optimize the above function. Readers are referred to \cite{McHutchon2013Differentiating} on how to differentiate GPs with a squared exponential co-variance function. In this work $\partial {Q}_{UCB}(s,a) / \partial a$ is numerically approximated with the Taylor series.

We assume each action dimension $a_j$ to be restricted to a symmetric range $[-\xi_j,\xi_j]$. 
In mathematical terms, this can be formulated as:
\begin{equation}  
\label{eqBoundMT}
a_{j} =\xi_j \frac{1 - \exp(-\alpha_j)}{1 + \exp(-\alpha_j)}.
\end{equation}
The unbounded action parameter $\alpha_j$ is found such as to minimize the given cost function. 
In order to find the optimal action we apply the Resilient backpropagation (Rprop) method (see \cite{ghadirzadeh2016self} for more details).

\begin{algorithm}[t!]

\SetKwInOut{Input}{input}
\caption{Gaussian Process Q-learning.}
\Input{Initial policy $\pi$}
Initialize $Q(s,a)$\;
\For{each iteration}{
	Run $\pi$ and collect $[s_i, a_i,s_{i+1}]_{i = 1, ..., T}$ on the real robot\;
    Train $\mathscr{F}$ with $[s_i, a_i,(s_{i+1} - s_i)]_{i = 1, ..., T-1}$\; 		
    \For{each episode}{
    sample $s \sim p(s_0)$\;
    $\pi' \leftarrow \argmin_{a} Q(s, a)$\;
    \For{each step in episode}{
    	choose $\epsilon$-greedy action $a$ w.r.t. $\pi'$\;
        predict the next state $s'$\;
    	update $Q(s,a)$ according to Eq.~\ref{eq:Qupdate}\;
        $s \leftarrow s'$ \;        
	  }
      }
$\pi \leftarrow \argmin_a Q_{UCB}(s,a)$\;
} 
\label{alg:Qlearning}
\end{algorithm}

\subsubsection{Learning the action-value function}
As explained earlier, training of the action-value function is based on the data provided by the forward model. Alg.\ \ref{alg:Qlearning} summarizes how this method works. The Q-function is initialized with the immediate predicted cost for a set of state-action pairs. In each iteration, the policy $\pi$, found according to Eq.~\ref{eq:policy}, is applied to the robot and generates a state-action trajectory. This trajectory is used to train the forward model. 

An episode corresponds to updating the Q-function with a simulated trajectory generated by the forward model. The $\epsilon$-greedy policy $\pi'$ w.r.t. $Q(s,a)$ is applied to generate the simulated samples. This is an important feature of model-based approaches that we can explore policies which are not possible to run on the real robot. Furthermore, probabilistic representation of the Q-learning rule (Eq. ~\ref{eq:Qupdate}) allows updating the Q-function even for uncertain state-transition data. Also, this approach does not suffer from the error-accumulation problem, which is an issue for model-based approaches, 
since Q-learning only requires the data for a single state-transition to make the update in each step. 

In the next section, our experimental results will be presented.

\begin{figure}[b!]
\begin{center}
\includegraphics[width=0.49\textwidth ]{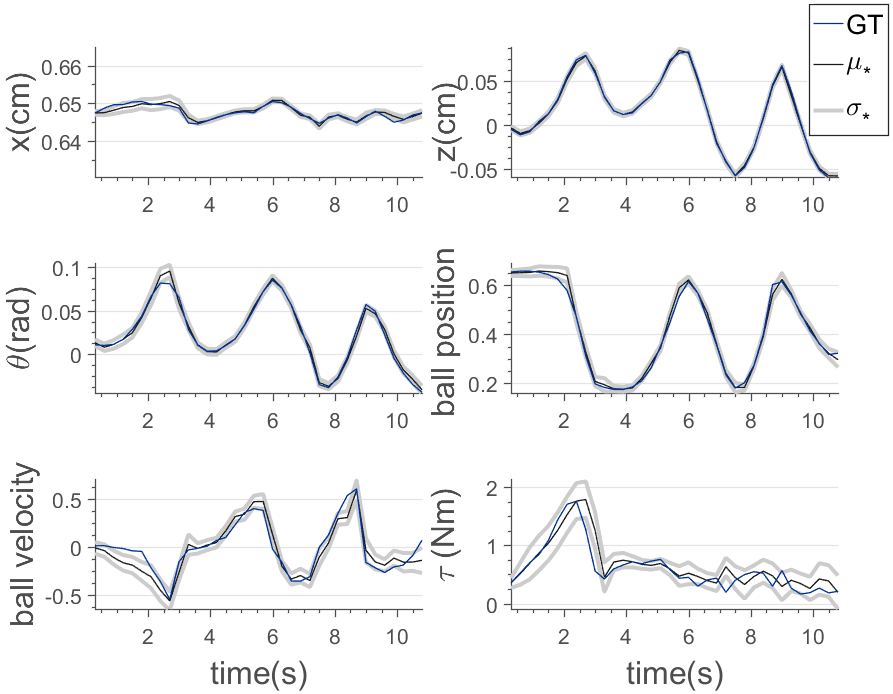}
\caption{Forward model predictions for a novel trajectory is compared against the ground truth. The prediction standard deviation is indicated by gray.}
\label{fig:forwardModel}
\end{center}
\end{figure}

\section{EXPERIMENTS}

In this section, we present the performance of our framework to learn joint controlling in a collaborative setting.  
The experimental setup is introduced in Sec.~\ref{sec:exp_setup}.
We present the prediction performance of the trained forward model and how human behavior modeling is achieved in Sec.~\ref{sec:exp_fm}.
Finally, the results on collaborative control task is presented in Sec.~\ref{sec:exp_ball_control}.

\subsection{Experimental setup}
\label{sec:exp_setup}
The PR2 robot shown in Fig.~\ref{fig:setup} is used to perform the experiments. 
In all experiments, the robot holds the plank with the left hand and a human collaborator holds the other side of the plank with both hands. 
The task is to control the position of a ball that can freely roll on the plank. 
The robot arm is a 7-degree-of-freedom manipulator and each joint is controlled by a PID velocity controller. 
The wrist of the robot is equipped with a 6-dimensional force/torque sensor. 
The vision system returns a noisy measurement of the ball position and velocity in a scaled unit ranged $[0, 1]$ at 4Hz. 
The whole framework is implemented in C++ while Gaussian process hyperparameters are found with the GPML Matlab toolbox \cite{rasmussen2010gaussian}. 

The forward model is initially trained by $150$ samples corresponding roughly to one minute of operation. The forward model predicts the next state $S_{t+1} = [x_{t+1},z_{t+1},\theta_{t+1},d_{t+1}, \dot{d}_{t+1},\tau_{t+1}$] based on the current state-action pair $x_{t}= [s_{t}, a_{t}]$. 
The objective function determines the importance of different terms by the diagonal weight matrix $W_c$. In the first setup, we penalize ball positioning error and ball velocity by setting the corresponding weights to $[1.0, 0.2]$ and the remaining ones to zero. In the second setting, we additionally set the interaction force weight to $0.01$ to analyze its effect on the smoothness of the interaction.
Actions, defined as the end-effector Cartesian velocity, are limited by $\xi_1 = \xi_2 = 0.1m$/$s$ in the X-Z plane and $\xi_3 = 0.2 rad$/$s$ for the pitch angular velocity.
The discount factor introduced by Eq.~\ref{eq:Qupdate} is set to $\gamma = 0.2$, and the Bayesian optimization weight $\delta$ in Eq.~\ref{eq:costPi} is set to $-0.5$. 

\subsection{Forward model learning}
\label{sec:exp_fm}
The forward model predicts state transitions resulting from the robot's actions. We present our results to demonstrate the forward model capability to predict such state-transitions for unseen state-action trajectories. Furthermore, using GP's automatic relevance determination, we show predicted relevant sensorimotor dimensions for each output of the forward model. Finally, we experimentally show how human behavior can be represented by the learned sensorimotor contingencies of the robot. 
 
Forward model state-transition prediction for a novel action-state trajectory is presented in Fig.~\ref{fig:forwardModel}. 
The next state is predicted with a normal distribution $\mathcal{N}(\mu_*, \sigma_*^2)$, where $\mu_*$ and $\sigma_*^2$ are the GP posterior mean and variance. 
As shown in the figure, the forward model predicts $x$, $z$, $\theta$ and the ball position with a good precision. However, the ball velocity and the interaction force are predicted less precisely. The ball velocity is less predictable because of the noisy vision system. The interaction force is hard to predict due to stochasticity inherent in the human actions. However, the learning rule in Eq.~\ref{eq:Qupdate} can handle this issue, since the prediction uncertainty indicated by the posterior standard deviation represents the true bound for the prediction error. Therefore, each update in the Q-function is taking place with correct information. 

\begin{figure}[t!]
\begin{center}
\includegraphics[width=0.3\textwidth ]{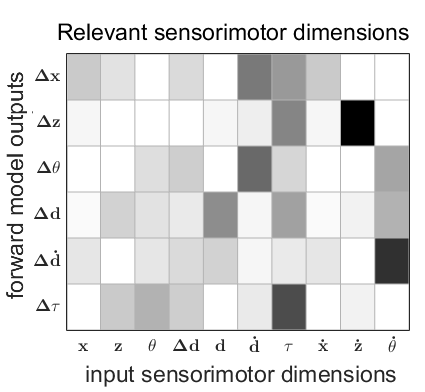}
\caption{The relevant inputs for the different dimensions of the forward model. The darker the indicating color, the more important is this input dimension to the respective output.}
\vspace{-8 mm}
\label{fig:relDim}
\end{center}
\end{figure}

Next, we discuss relevance determination for each sensorimotor dimension as the input of the forward model.  
The relevance of an input dimension is indicated by the value of the $\lambda$s in Eq.~\ref{eq:covSEard}. 
The higher the value is, the more relevant is the corresponding dimension to the regression output. 
Fig.~\ref{fig:relDim} demonstrates a color-coded representation of the $\lambda$s for each dimension of the forward model, and therefore shows the relevance of the corresponding sensorimotor dimension. 
As an example on how to interpret the table, one can look at the $\Delta z$ output. The two input dimensions, $\tau$ and $\dot{z}$, have the darkest values indicating their importance for predicting the next value of the $z$ position.

As demonstrated by the Fig.~\ref{fig:relDim}, $\theta$ is highly correlated with the ball velocity and the commanded angular velocity. 
One explanation for why $\theta$ and $\dot{d}$ are correlated is that the partners tilt the board for both cases when the ball velocity is too high or too low. 
Hence one clue to predict $\theta$ is to observe the ball velocity. The ball position is mostly governed by the current ball position, the interaction force and the commanded angular velocity. Its dependency on the ball velocity might have been eliminated because of the fact that the ball velocity itself is highly correlated with the commanded angular velocity. 
The interaction force $\tau$ beside being dependent on its old value, it is also dependent on $\theta$, $z$ and the ball distance to the target position. 
All the last three dependencies explain human behavior in different conditions and therefore are required to predict the interaction force. 

\begin{figure}
    \centering
    \begin{subfigure}[b]{0.3\textwidth}
        \includegraphics[width=\textwidth]{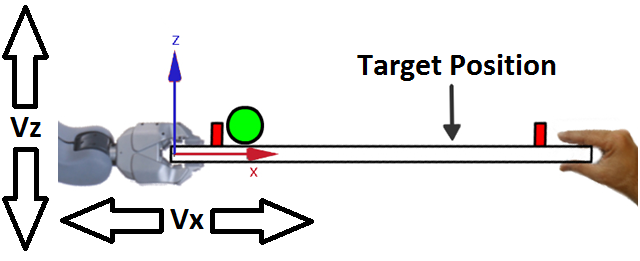}
        \caption{The simulated robot actions and the configuration of the system. }
        \label{fig:setup2}
    \end{subfigure}
     ~ 
    \begin{subfigure}[b]{0.35\textwidth}
        \includegraphics[width=\textwidth]{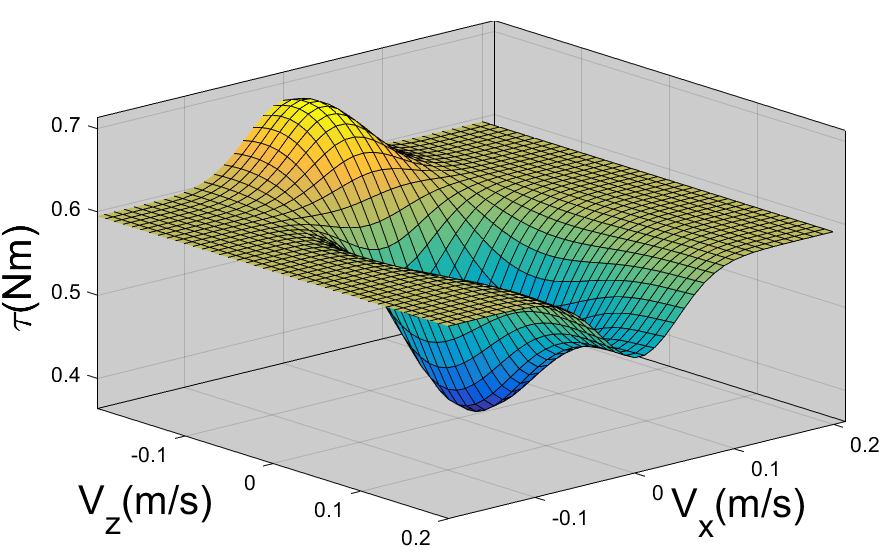}
        \caption{Forward model predictions of the interaction force for the given range of the actions}
        \label{fig:force}
    \end{subfigure}   
    \caption{An illustration of how the robot can model human behavior based on its own sensorimotor modeling.}
	\label{fig:human_pred}
\end{figure}

Finally, we explain how the robot's sensorimotor model of the interaction can represent human behavior. The human partner has an important share for the total interaction force measured at the robot's wrist. The measured torque can be increased in the case that the human resists against the robot's motion plan or it can be decreased when, otherwise, the human supports the motion. Therefore, since the robot can predict the interaction force for future times, it can indirectly predict the partner's actions. Fig.~\ref{fig:human_pred} shows an example interaction force predictions for a range of robot's actions. As it is depicted in Fig.~\ref{fig:setup2}, the robot is pushing and pulling the plank horizontally and vertically.  At Fig.~\ref{fig:force}, the interaction force is predicted based on the forward model for the different actions. As it is demonstrated, horizontal pulling or pushing both increases the predicted interaction force since the human partner will resist against those motions. 
Lifting the board vertically generates less interaction force considering the ball target position shown in the figure, which agrees the forward model predictions. 
On the other hand, lowering the board horizontally generates more torque, since the human partner tries to tilt the board in the opposite direction and again this is well-predicted by the forward model. Similar arguments can be made for other sensory observations which are consequences of both partners' actions.           

\begin{figure}
\begin{center}
\includegraphics[width=0.4\textwidth ]{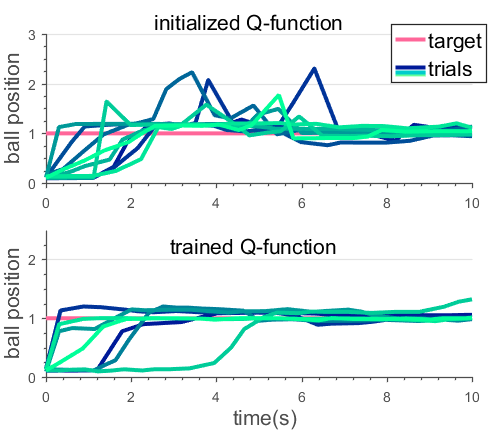}\caption{
The step response of controlling the ball position for different trials.}
\vspace{-8 mm}
\label{fig:qcompare}
\end{center}
\end{figure}

\begin{figure*}[t!]
	\vspace{0.3cm}
    \centering
    \includegraphics[width=1\textwidth]{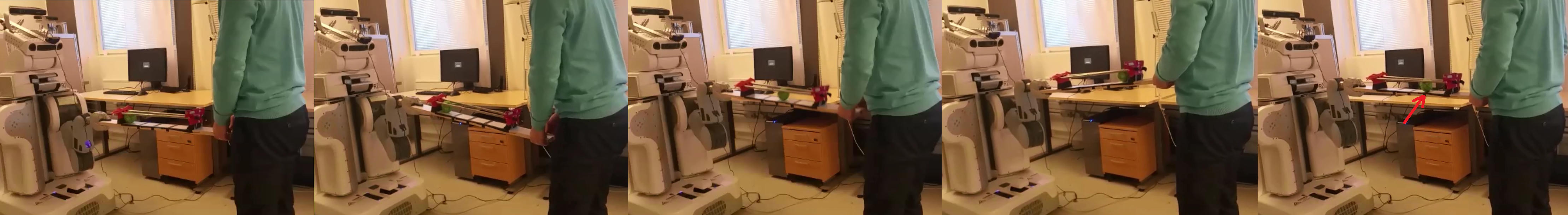}
    \caption{
	Snapshots of the collaboratively ball positioning with the target ball position shown by the red arrow.  
	}
    \label{fig:control}
\end{figure*} 

\subsection{Collaborative ball control}
\label{sec:exp_ball_control}
In this part, we present the experimental result for the collaborative control task. 
Fig.~\ref{fig:control} shows some snapshots of the joint human-robot ball control and Fig.~\ref{fig:qcompare} illustrates the results quantitatively for the two cases 1) the Q-function is just initialized with the forward model data and 2) the Q-function is repeatedly trained according to Eq.~\ref{eq:Qupdate}.  
For each case a number of trial is presented. 
It is apparent that the overshoot of the ball position improves considerably when the Q-function is trained iteratively.

\begin{figure}[t!]
\begin{center}
\includegraphics[width=0.45\textwidth ]{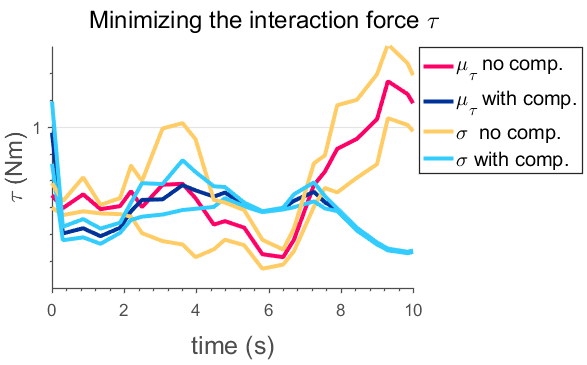}
\caption{The average interaction force of several trials for the two policies with and without interaction force compensation.}
\vspace{-8 mm}
\label{fig:mintau}
\end{center}
\end{figure}

Finally, we present how the workload of the human, measured as the interaction force $\tau$, can be minimized by including it into the cost function.  
Fig.~\ref{fig:mintau} demonstrates the average interaction force over 10 trials for each of the two cases, with and without the interaction force compensation. 
Firstly, the measured interaction force has a lower variance over the trials for the with-compensation case. This implies the robot will behave more similar w.r.t. the interaction force sensed by the human partner compared to the case which the only aim is to control the ball position.   
Secondly, the average interaction force is lowered for the beginning and the end of the trials. This implies the robot starts more smoothly and avoids vibrating at the end.

\section{CONCLUSIONS AND FUTURE WORK}
In this work, we have presented a sensorimotor reinforcement learning framework to solve a pHRI task. The main contribution of this work is the probabilistic sensorimotor formulation of human behavior. We avoid high-level modeling of human actions by directly grounding the task in the sensorimotor space of the robot. As an example, we have shown in the experiments that the model learns the relation between interaction force (which is mainly caused by the human) and its effect on the ball position without any explicit modeling. 

We use Gaussian processes for sensorimotor modeling that also capture the underlying uncertainty in the system. 
Bayesian optimization is applied to operate safely even under uncertainties and model imperfections.   

Data-efficiency is a highly important characteristic in designing this framework. In fact, collecting many training samples, as it is required by traditional reinforcement learning methods, might be infeasible generally in pHRI tasks. 
We have observed experimentally that Gaussian processes predict sufficiently good with training data recorded roughly in one minute of operation. 
Furthermore, the Q-function is updated with the simulated data made by the forward model instead of direct data-queries to the robot. 
These two features of the framework result in a data-efficient learning suitable for pHRI tasks.  

Equal role sharing is achieved by defining a cost function for the robot similar to the one of the human partner. 
Furthermore, the two partners mutually predict each other and therefore avoiding conflicts, while both trying to exert the own motion plan. 

This work has demonstrated how a physical interaction for a dyad can be grounded in the raw sensorimotor observations of each.   
As our future work, we will study similar interactions with the robot equipped with richer sensory observations and will investigate the effect of this on the level of coupling between the two partners.

\addtolength{\textheight}{-12cm}   




\section*{ACKNOWLEDGMENT}
This work was supported by the EU through the project socSMCs (H2020-FETPROACT-2014) and the Swedish Research Council.

\bibliographystyle{IEEEtran}
\bibliography{root}

\end{document}